\documentclass[a4paper, 10pt, conference]{ieeeconf}      % Use this line for a4
\usepackage{FG2020}

\FGfinalcopy % *** Uncomment this line for the final submission

\IEEEoverridecommandlockouts                              % This command is only
\usepackage{graphics} % for pdf, bitmapped graphics files
\usepackage{epsfig} % for postscript graphics files
\usepackage{mathptmx} % assumes new font selection scheme installed
\usepackage{times} % assumes new font selection scheme installed
\usepackage{amsmath} % assumes amsmath package installed
\usepackage{amssymb}  % assumes amsmath package installed
\usepackage{cite}
\usepackage{amsfonts}

\def\etal{\emph{et al}.}
\def\eg{\emph{e.g}.}

\def\FGPaperID{0118} % *** Enter the FG2020 Paper ID here

\title{\LARGE \bf
	Landmarks-assisted Collaborative Deep Framework for \\Automatic 4D Facial Expression Recognition
}

\author{\parbox{16cm}{\centering
		{\large Muzammil Behzad, Nhat Vo, Xiaobai Li and Guoying Zhao}\\
		{\normalsize
			Center for Machine Vision and Signal Analysis (CMVS), University of Oulu, Finland\\
			Email: \{muzammil.behzad, nhat.vo, xiaobai.li, guoying.zhao\}@oulu.fi
		}}
	\thanks{This work was supported by Infotech Oulu, the National Natural Science Foundation of China (No. 61772419), Tekes Fidipro Program (No. 1849/31/2015), Business Finland Project (No. 3116/31/2017), and Academy of Finland. As well, the authors wish to acknowledge CSC – IT Center for Science, Finland, for computational resources.}% <-this % stops a space
}

\begin{document}
	
	\ifFGfinal
	\thispagestyle{empty}
	\pagestyle{empty}
	\else
	\author{Anonymous FG2020 submission\\ Paper ID \FGPaperID \\}
	\pagestyle{plain}
	\fi
	\maketitle

	%%%%%%%%%%%%%%%%%%%%%%%%%%%%%%%%%%%%%%%%%%%%%%%%%%%%%%%%%%%%%%%%%%%%%%%%%%%%%%%%
	\begin{abstract}
		We propose a novel landmarks-assisted collaborative end-to-end deep framework for automatic 4D FER. Using 4D face scan data, we calculate its various geometrical images, and afterwards use rank pooling to generate their dynamic images encapsulating important facial muscle movements over time. As well, the given 3D landmarks are projected on a 2D plane as binary images and convolutional layers are used to extract sequences of feature vectors for every landmark video. During the training stage, the dynamic images are used to train an end-to-end deep network, while the feature vectors of landmark images are used train a long short-term memory (LSTM) network. The finally improved set of expression predictions are obtained when the dynamic and landmark images collaborate over multi-views using the proposed deep framework. Performance results obtained from extensive experimentation on the widely-adopted BU-4DFE database under globally used settings prove that our proposed collaborative framework outperforms the state-of-the-art 4D FER methods and reach a promising classification accuracy of 96.7$\%$ demonstrating its effectiveness.
	\end{abstract}

	%%%%%%%%%%%%%%%%%%%%%%%%%%%%%%%%%%%%%%%%%%%%%%%%%%%%%%%%%%%%%%%%%%%%%%%%%%%%%%%%
	\section{INTRODUCTION}
	\label{sec:intro}
	Facial expressions (FEs) are important cues in understanding human emotions during their social communication. To better facilitate the understanding and analysis of such FEs, many researchers proposed facial expression recognition (FER) systems using the state-of-the-art computer vision based human-machine interaction methods. As a result, the research community has witnessed tremendous surge in FER systems towards potentially significant application areas like psychology, security, education, bio-medical and computing technology. The study concluded by Ekman and Friesen~\cite{ekman1971constants} serves as a pioneer contribution in this field dating back to 1970s. This work presented the six globally-adopted human facial expressions which are happiness, anger, sadness, fear, disgust and surprise.
	
	In the past few years, several machine learning methods were presented for recognizing facial expressions with the help of static or dynamic 2D images. Despite promising contributions, however, emotion recognition still remains a challenging problem due to the sensitivity of 2D images towards lighting conditions, pose variations and occlusions~\cite{fasel2003automatic}. This is why 2D based methods are not fully stable and could not potentially contribute to real-world applications beyond a certain point. As a rival, 3D point clouds rescued and motivated novel FER directions via trending high-resolution and high-speed 3D data acquisition equipment. Although the data processing becomes complex, the significantly increased amount of data in terms of the facial deformation patterns over the depth axis considerably help the deep learning models to learn patterns effectively for automatic~FER.
	
	Importantly, although each facial expression is in fact a combination of different muscle movements ordered in a particular way, consequently triggering facial deformations~\cite{5771466}, such compact cues are better captured in geometrical domain \cite{7457243}. This is why the 3D face scans are quite convenient in representing such deformations, and therefore, predicting the emotions. The collection of different large-size and complex databases with various terabytes of data has supported such research on FER using 3D face scans. In this regard, the release of commonly known BU-3DFE \cite{yin20063d} and Bosphorus~\cite{savran2008bosphorus} served as one of the pioneering datasets for investigating FER via static 3D data. The dynamic 3D data (referred as 4D), such as the BU-4DFE dataset \cite{4813324}, allows to perform 4D FER by fetching facial deformations both in the depth geometry and over time.
	
	Contrary to the 3D FER methods which only rely on the static data at hand \cite{5597896,6460694,zhen2015muscular,li2015efficient}, 4D face scans enable deep networks to learn effectively for better analyzing and predicting facial expressions. In this regard, Sun \etal~\cite{Sun:2010:TVF:1820799.1820803} and~Yin~\etal~\cite{4813324} proposed to work around Hidden Markov Models (HMM) to learn the facial muscle patterns over time. Similarly to benefit from local facial patterns, Drira~and~Amor~\cite{6460329, amor20144} introduced a deformation vector field mainly based on Riemannian analysis and combined with random forest. In another attempt, Sandbach \etal~\cite{sandbach2012recognition} represented 3D frames and its neighbors as Free-Form Deformation (FFD) and subsequently used HMM and GentleBoost as classifiers. Using the traditional Support Vector Machine (SVM), the authors represented geometrical coordinates and its normal as feature vectors \cite{FANG2012738}, and as dynamic Local Binary Patterns (LBP) in an extended work \cite{6130440}. Likewise, a spatio-temporal LBP-based feature was proposed in \cite{6553746} to extract features from polar angles and curvatures.
	
	Yao \etal~\cite{Yao:2018:TGS:3190503.3131345} applied Multiple Kernel Learning (MKL) by using the scattering operator \cite{bruna2013invariant} on 4D face scans to produce effective feature representations. In a similar attempt to recognize FEs, statistical shape model with global and local constraints were proposed in \cite{fabiano2018spontaneous}. The authors claimed that local shape index and global face shape can together help build a desirable FER system. A much effective deep network was proposed by Li \etal~\cite{8373807} to automate 4D FER using a dynamic geometrical image network. In this work, geometrical images were generated after the differential quantities were estimated. The final prediction step involved fusing the predicted scores from different geometrical images. Bejaoui~\etal~\cite{Bejaoui2019} recently proposed a sparse coding-based representation of LBP difference. They extracted a unified set of geometric and appearance features via Mesh-Local Binary Pattern Difference (mesh-LBPD), combined them into a compact representation via covariance matrices, and then applied sparse coding for effective 3D/4D~FER.
	
	Despite these attempts to automate FER via 4D data, we believe that the facial deformations should be appropriately extracted from the spatio-temporal 4D data for better network learning rather than simply tuning multiple parts of a deep network. Consequently, in this paper, we aim to fetch such deformations jointly from multi-views and some geometrical domains to propose a landmarks-assisted collaborative end-to-end deep framework for automatic 4D FER (LC4D). We project every 3D face scan to extract various geometrical images such as depth images and texture images. For a robust representation of the facial features, we aim to extract the features across various multi-views. To encode the muscle movements of different expressions from the temporal domain, we apply rank pooling to compute the dynamic images of the 4D data. As well, we represent the stored facial movements from the 3D landmarks by projecting them on a 2D plane. Afterwards, activations of these sequences of landmarks are computed using convolution layers, which are then trained on an LSTM network. Using decision-level fusion, the landmarks then collaborate with the dynamic images trained over another deep network for an improved FER by highlighting the correct classification probabilities. To our best knowledge, this is the first landmarks-assisted collaborative deep framework for 4D FER using multi-views.
	
	The rest of the paper is sectioned as follows: Section \ref{sec:proposedmethod} explains our proposed LC4D method for 4D FER in detail. In Section \ref{sec:results}, the experimental results of our framework are reported. Finally, Section \ref{sec:conclusions} concludes the paper.

	%%%%%%%%%%%%%%%%%%%%%%%%%%%%%%%%%%%%%%%%%%%%%%%%%%%%%%%%%%%%%%%%%%%%%%%%%%%%%%%%	
	\section{Proposed Automatic 4D FER Method}
	\label{sec:proposedmethod}
	In this section, we discuss in detail the working mechanism of our proposed LC4D deep framework. First, we explain the filtering step to remove the unwanted and noisy mesh components in the given 3D point clouds. Then, we discuss the computation of different geometrical images in multi-views. Third, we elaborate the landmarks-assisted mechanism. Finally, the collaborative scheme for 4D FER is presented. An overview of our method is shown in Fig. 1.
	
	\subsection{Pre-processing}
	The 3D point clouds from BU-4DFE contain noise and unwanted components like outliers, hair, and non-facial regions, which causes a deep model incapable of learning effectively, and therefore, should be filtered out.
	\begin{figure*}[t!]
		\includegraphics[width=\linewidth]{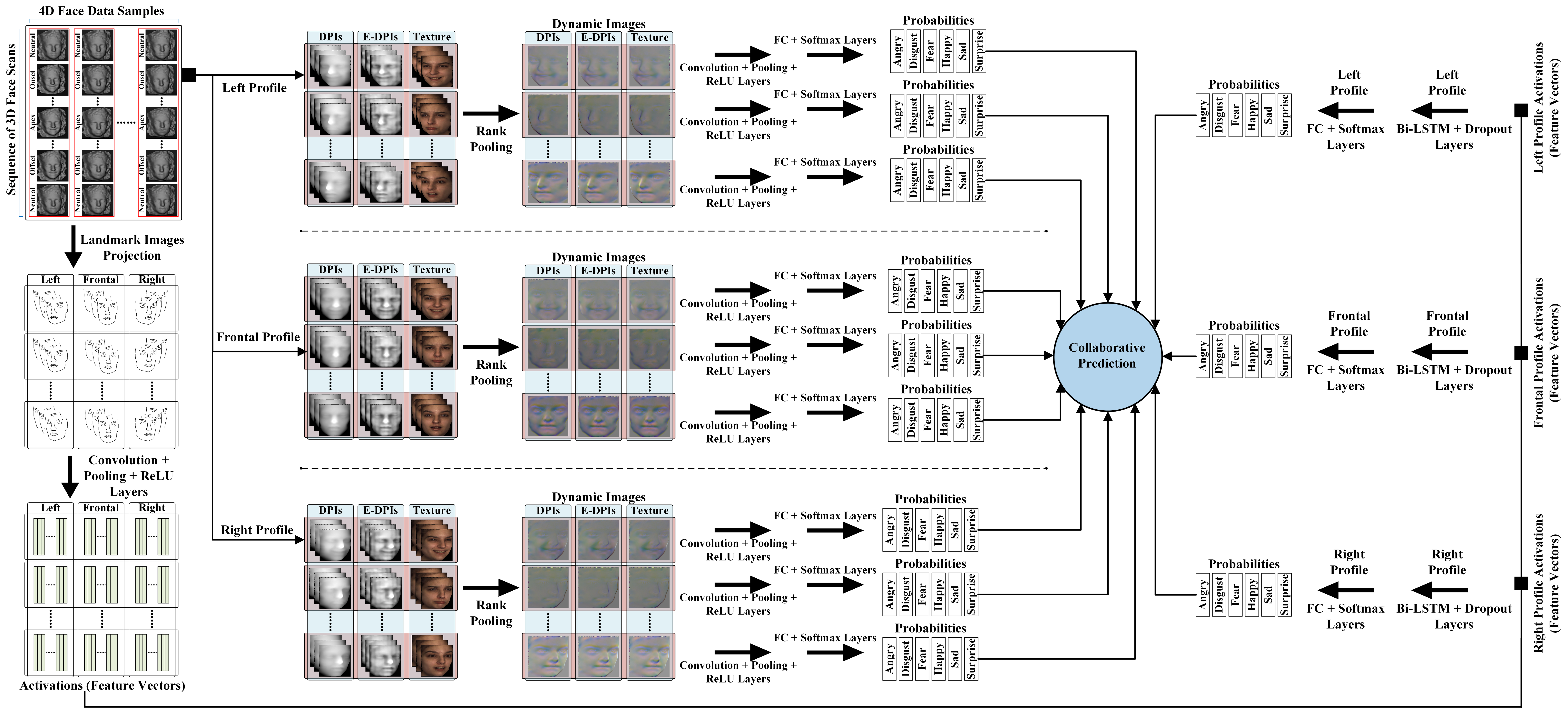}
		\centering
		\caption{The proposed LC4D method for 4D FER.}
		\label{fig:4DFER_floechart}
	\end{figure*}
	
	Subsequently, we pre-process each 3D point cloud from $N$ 4D samples individually to combat such outliers. We define
	\begin{equation}
	\label{eq:1}
	I^{4D} = \{I_{nt}^{3D}\},  \text{ } \forall t = \{1,2,...,T_n\} \text{ and }\forall n = \{1,2,...,N\},
	\end{equation}
	where $I^{4D}$ refers to 4D samples, and $I_{nt}^{3D}$ is the $n$th point cloud at $t$th frame. Note that (\ref{eq:1})~$\Rightarrow |I^{4D}| = N \text{, and } |I_{nt}^{3D}| = T_n$. The mesh with $M$ vertices can be denoted as
	\begin{equation}
	\label{eq:2}
	\textbf{m} = [\textbf{v}_1^T,...,\textbf{v}_M^T]^T = [x_1,y_1,z_1,...,x_M,y_M,z_M]^T,
	\end{equation}
	where $\textbf{v}_j = [x_j,y_j,z_j]^T$ are coordinates of the $j$th vertex, and $\textbf{m}_{t}$ is a mesh at $t$th frame such that $\textbf{m}_{n} = \{\textbf{m}_{nt}\} \forall n$. To crop the unwanted regions via given landmarks, we remove everything beyond the facial-border landmarks. For the head-hair, we trim regions above the forehead by using a threshold which is a fractional distance from eyebrows to the tip of nose. This is denoted as
	\begin{equation}
	\label{eq:3}
	\overline{I_{nt}^{3D}} = \eta_c(I_{nt}^{3D}),
	\end{equation}
	where $\overline{I_{nt}^{3D}}$ is the cropped face and $\eta_c(.)$ is the cropping operation. This also updates the \textit{mesh-space} in (\ref{eq:2}) as follows:
	\begin{equation}
	\label{eq:4}
	\overline{\textbf{m}} = \eta_c(\textbf{m}) = [\textbf{v}_1^T,...,\textbf{v}_{\overline{M}}^T]^T = [x_1,y_1,z_1,...,x_{\overline{M}},y_{\overline{M}},z_{\overline{M}}]^T,
	\end{equation}
	where $\overline{\textbf{m}}$ is the set of updated vertices such that $\overline{\textbf{m}} \subseteq \textbf{m}$ and $\overline{M} \le M$. The filtering steps in (\ref{eq:3}) and (\ref{eq:4}) remove all the outliers that would potentially disturb the training process.	
	
	\subsection{Geometrical Images over Multi-views}	
	Geometrical images provide efficient feature mapping from 3D to 2D \cite{7163090}. Therefore, after pre-processing, we compute the depth images (DPI) as $f_D:~\overline{I^{3D}}~\rightarrow~I_D$, and the texture images as $f_T: \overline{I^{3D}} \rightarrow I_T$ from the filtered meshes via 3D to 2D rendering, where $f_D$ and $f_T$ denote the function mapping to depth image $I_D \in \mathbb{R}^{K^2}$, and texture image $I_T \in \mathbb{R}^{K^2}$, respectively, where $K^2$ is the number of pixels. For sharp facial details, the contrast-limited adaptive histogram equalization is applied on DPIs to get enhanced-depth images (E-DPIs) as $I_{ED} = \eta_s(I_D)$, where $\eta_s(.)$ refers to the sharpening operator. Importantly, we also generate these images in alignment profiles from right-to-frontal-to-left for an effective collaboration at a later stage in the network.
	
	Once we compute the pre-processed images from different domains (\eg, texture and depth) over multi-views, as depicted in Fig. \ref{fig:4DFER_floechart}, the next step is to fetch and represent the variations of the facial deformations from the temporal domain. One optimal choice is to perform rank pooling over the projected 2D sequences of geometrical images to obtain their dynamic images \cite{bilen2018action}. Consequently, we perform rank pooling to represent the temporal dynamics of entire videos into a single RGB image. We compute the dynamic images for all 4D samples using the projected sequences over multi-views. All the extracted cross-domain dynamic images, shown in Fig. \ref{fig:4DFER_floechart}, incorporate the spatio-temporal patterns effectively which is favorable for training a deep network. It is worth mentioning that this idea is different from the one proposed in~\cite{8373807}. This is because they used geometrical images independently, while we use the dynamic images of different domains to collaborate over multi-views. More importantly, as explained in the subsequent section, we also deploy a deep landmarks-based network ultimately improving the classification scores for 4D FER, which, to the best of our knowledge, has never been reported in the literature before.
	
	\subsection{Landmarks-assisted Learning}
	Since landmarks encode key facial points that potentially represent an expressed emotion, our landmarks-assisted approach significantly benefits the proposed framework for 4D FER. To do so, we first similarly pre-process all the given 83 landmarks for each frame. Then, we project them on a 2D plane across various multi-views for representing the facial deformations stored in landmarks over time as binary images. With the sequences of such projected landmark images as input, we extract its activations by using convolution and pooling layers. For this purpose specifically, we used a pre-trained GoogLeNet \cite{szegedy2015going} to convert the videos of landmark images into sequences of feature vectors containing appropriate feature presentation of each frame as a vector. 
	
	Finally, we create a long short-term memory (LSTM) network with a sequence input layer, Bi-LSTM layer with 2000 hidden units, 50$\%$ dropout layer followed by FC, Softmax and classification layer. This is first trained on the sequences of these feature vectors, and is then used to predict the expressions. Note that the parameterized model in \cite{8756614} just use landmarks to train and then predict expressions in a straight forward manner. Conversely, we propose a deep framework where the extracted sequences of feature vectors of the projected landmark images across multi-views are used to train an LSTM network first, and are then used in collaboration with the rest of the framework for an improved expression prediction as outlined in the next section.	
	
	\subsection{Collaborative Prediction}
	\label{subsec:Collaborative_Prediction}
	For improved predictions, a final collaborative step is performed to tailor the voting scores of different expressions using various collaborative elements. Specifically, since our network is trained after multiple Convolution+Pooling+ReLU layers, we jointly utilize the expression probabilities from different geometrical domains over multi-views. While doing so, the contributions from landmarks-assisted network are equally respected to have updated and much improved predictions. This is because even though the patterns from dynamic images already discriminate different expressions, the landmarks-assisted approach via multi-view further enhances the likelihood of correct predictions.	For six expressions, the predicted and collaboratively-updated probabilities, respectively, are as follows:
	\begin{equation}
	\label{eq:6}
	C = [\textbf{c}_1^T,...,\textbf{c}_N^T]^T, \text{ and}
	\end{equation}
	\begin{equation}
	\label{eq:7}
	C(n,l) = \frac{1}{|\Theta|}\sum_{\forall \theta \in \Theta} [C_{DI}(n,l_\theta) + C_{LI}(n,l_\theta)], \text{ }\forall n,l.
	\end{equation}
	Here, $\textbf{c} = \{\rho_l\} \text{ for } l = \{1,...,6\},$ is the predicted probabilities of six expressions for $n$th sample, $\Theta$ is all view angles over which multi-views are collected, while the subscript \textit{DI} and \textit{LI} refers to dynamic and landmark images, respectively. The finally updated predictions $F(n)$ are computed as maximum of all the expression probabilities
	\begin{equation}
	\label{eq:8}
	F(n) = \text{max}\{C(n)\}, \text{ }\forall n = \{1,2,...,N\}.
	\end{equation}

	%%%%%%%%%%%%%%%%%%%%%%%%%%%%%%%%%%%%%%%%%%%%%%%%%%%%%%%%%%%%%%%%%%%%%%%%%%%%%%%%
	\section{Experimental Results and Analysis}
	\label{sec:results}
	With the extensive training, we evaluate our experimental results to analyze the improvement in prediction performance based on our proposed LC4D method for FER. We opted the generally used BU-4DFE dataset for our experiments. This dataset contains 58 females and 43 males (a total of 101 subjects) each having all six human facial expressions, i.e., happiness, anger, sadness, fear, disgust and surprise. Every expression of a subject contains dynamic 3D data of raw face scans with a frame rate of 25 frames per second (fps) lasting approximately 3 to 4 seconds.
	
	For having a fair comparison of results with other state-of-the-art methods, we follow \cite{8373807} and choose similar experimental settings. In particular, we employ a 10-fold subject-independent cross-validation (10-CV), and use a 60-20-20 split of the data for training, validation and testing, respectively over five iterations. Instead of using key-frames~\cite{Yao:2018:TGS:3190503.3131345} or employing sliding windows \cite{sandbach2012recognition}, we use entire 3D sequences. For the training using dynamic images, we use the pre-trained VGGNet \cite{Parkhi15} as the deep network, and therefore resize our images to $224 \times 224$. For the landmarks, we use the pre-trained GoogLeNet to extract feature vectors from activations, and then train an LSTM network from scratch on the extracted activations. All of our experiments are carried out on a GP100GL GPU (Tesla P100-PCIE), and the overall training time takes approximately one day.
	
\begin{figure}[b!]
		\includegraphics[width=\linewidth]{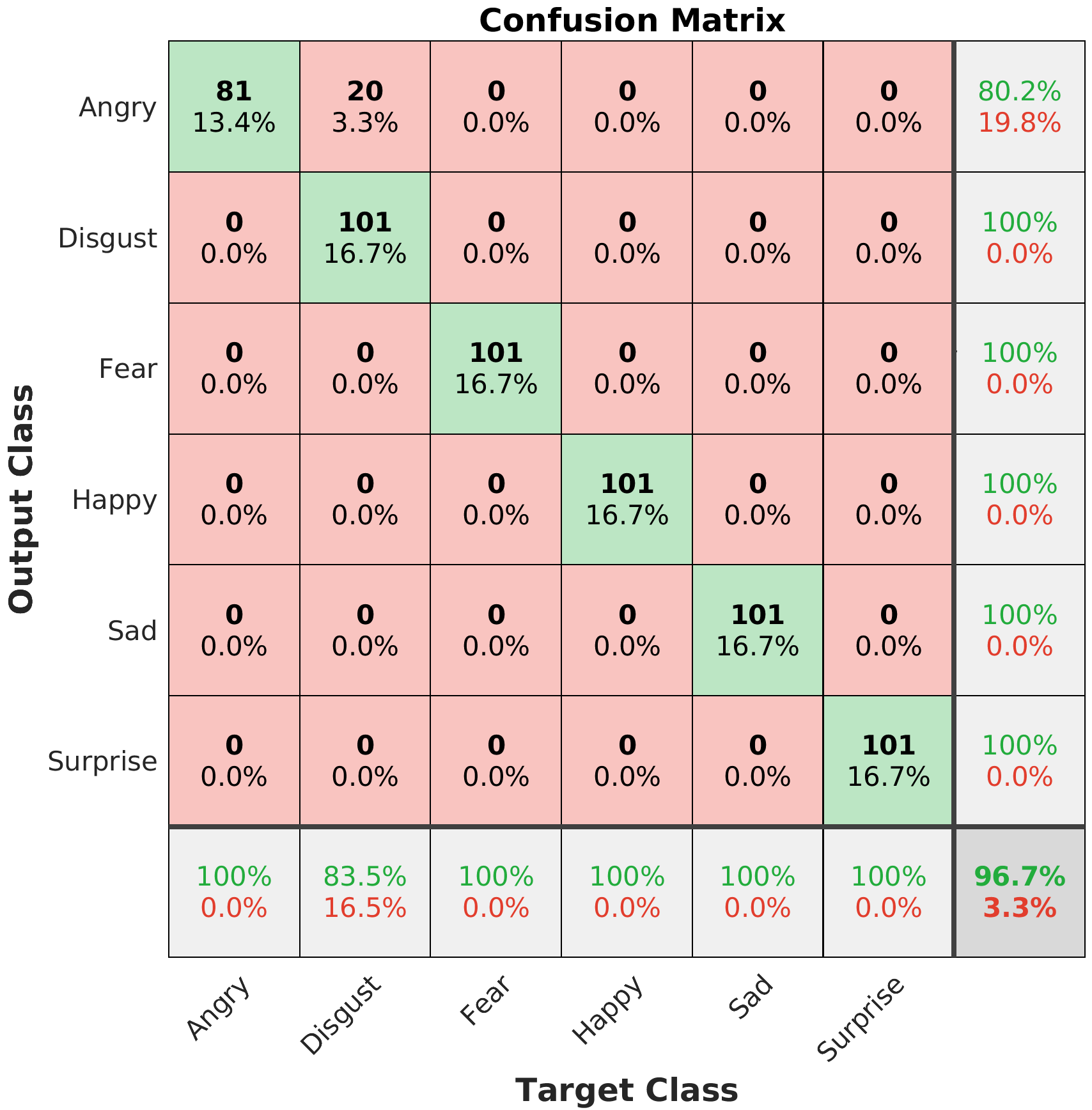}
		\centering
		\caption{Confusion matrix of predicting expressions on BU-4DFE database.}
		\label{fig:4DFER_CM}
	\end{figure}

	We show in Table \ref{table:4DFERLocalresults} the extensive comparisons of the classification accuracies calculated at various collaboration stages of our proposed method. As shown, the multi-views significantly help in revealing potential patterns and assist both landmarks as well as the dynamic images for better learning. It can be seen that promising results are achieved when all collaborators help in prediction. Similarly, we also show the confusion matrix of our experiments in Fig. \ref{fig:4DFER_CM}. Despite angry and disgust being some error cases due to their similarities, Fig. \ref{fig:4DFER_CM} indicates that our method correctly predicts emotions most of the time showing its effectiveness.

	\begin{table}[b!]
		\caption{FER accuracy reached on the BU-4DFE dataset. [LP = Left Profile, RP = Right Profile, FP = Frontal Profile]}
		\label{table:4DFERLocalresults}
		\begin{center}
			\begin{tabular}{||l|l|c||}
				\hline
				Collaborator(s) & Multi-view Profile(s) & FER Accuracy ($\%$) \\
				\hline\hline
				 & LP & 75.40 \\ \cline{2-3}
				 & FP & 78.70 \\ \cline{2-3}
				Landmark & RP & 77.60 \\ \cline{2-3}
				Images & RP + FP & 85.50 \\ \cline{2-3}
				 & LP + FP & 84.20 \\ \cline{2-3}
				 & RP + LP & 83.80 \\ \cline{2-3}
				 & RP + FP + LP & \textbf{88.80} \\
				\hline\hline
				 & LP & 78.30 \\ \cline{2-3}
				 & FP & 80.20 \\ \cline{2-3}
				Dynamic & RP & 79.20 \\ \cline{2-3}
				Images & RP + FP & 83.20 \\ \cline{2-3}
				 & LP + FP & 82.10 \\ \cline{2-3}
				 & RP + LP & 81.30 \\ \cline{2-3}
				 & RP + FP + LP & \textbf{84.70} \\
				\hline\hline
				 & LP & 83.40 \\ \cline{2-3}
				Landmark & FP & 91.40 \\ \cline{2-3}
				and & RP & 87.70 \\ \cline{2-3}
				Dynamic & RP + FP & 93.60 \\ \cline{2-3}
				Images & LP + FP & 92.10 \\ \cline{2-3}
				 & RP + LP & 88.80 \\ \cline{2-3}
				 & RP + FP + LP & \textbf{96.70} \\ \hline
			\end{tabular}
		\end{center}
	\end{table}
	
	Finally, in Table \ref{table:4DFERresults}, we present accuracies achieved on the BU-4DFE dataset by our proposed framework and several state-of-the-art methods \cite{sandbach2012recognition,6130440,7045888,Sun:2010:TVF:1820799.1820803,7457243,Yao:2018:TGS:3190503.3131345,FANG2012738,8373807,amor20144,8023848}. As shown in the table, our proposed method outperforms the existing methods while predicting the correct expression during classification. This is because of the extensively collaborative nature of our proposed framework in which the predictions are refined when the probability scores are updated from neighboring resources. The refinement in probabilities come from the fact that facial deformation patterns are well-captured in the geometric domain and its temporal movements are caught in the dynamic images. Importantly, the significant amount of assistance received from landmarks images also helped in making a reliable classification decision. Consequently, our LC4D framework reached an accuracy of $96.7\%$ for 4D~FER.
	\begin{table}[t!]
		\caption{Accuracy ($\%$) comparison with the state-of-the-art on the BU-4DFE dataset.}
		\label{table:4DFERresults}
		\begin{center}
			\begin{tabular}{||l|l|c||}
				\hline
				Method & Experimental Settings & Accuracy \\
				\hline\hline
				2012 - Sandbach \etal \cite{sandbach2012recognition} & 6-CV, Sliding window & 64.60 \\ \hline
				2011 - Fang \etal \cite{6130440} & 10-CV, Full sequence & 75.82 \\ \hline
				2015 - Xue \etal \cite{7045888} & 10-CV, Full sequence & 78.80 \\ \hline
				2010 - Sun \etal \cite{Sun:2010:TVF:1820799.1820803} & 10-CV, - & 83.70 \\ \hline
				2016 - Zhen \etal \cite{7457243} & 10-CV, Full sequence & 87.06 \\ \hline
				2018 - Yao \etal \cite{Yao:2018:TGS:3190503.3131345} & 10-CV, Key-frame & 87.61 \\ \hline
				2012 - Fang \etal \cite{FANG2012738} & 10-CV, - & 91.00 \\ \hline
				2018 - Li \etal \cite{8373807} & 10-CV, Full sequence & 92.22 \\ \hline
				2014 - Ben Amor \etal \cite{amor20144} & 10-CV, Full sequence & 93.21 \\ \hline
				2016 - Zhen \etal \cite{8023848} & 10-CV, Full sequence & 94.18 \\ \hline
				2019 - Bejaoui \etal \cite{Bejaoui2019} & 10-CV, Full sequence & 94.20 \\ \hline
				2018 - Zhen \etal \cite{8023848} & 10-CV, Key-frame & 95.13 \\ \hline
				\textbf{Ours} & 10-CV, Full sequence & \textbf{96.70}\\ \hline
			\end{tabular}
		\end{center}
	\end{table}

	%%%%%%%%%%%%%%%%%%%%%%%%%%%%%%%%%%%%%%%%%%%%%%%%%%%%%%%%%%%%%%%%%%%%%%%%%%%%%%%%	
	\section{Conclusions}
	\label{sec:conclusions}
	We proposed a 4D FER method via landmarks-assisted collaborative end-to-end deep framework. In this framework, different geometrical images were extracted first and were later used to capture facial movements over time in terms of compact dynamic images. Additionally, efficient feature representations were extracted from landmark images that were then used to train an LSTM network. An effective collaboration step performed over multi-views served as an added advantage of our deep framework. With a promising accuracy of $96.7\%$, our method outperformed the state-of-the-art 4D FER methods in terms of classification accuracy.

	%%%%%%%%%%%%%%%%%%%%%%%%%%%%%%%%%%%%%%%%%%%%%%%%%%%%%%%%%%%%%%%%%%%%%%%%%%%%%%%%
%	\section{ACKNOWLEDGMENTS}
%	
%	The authors gratefully acknowledge the contribution of reviewers' comments, etc. (if desired). Put sponsor acknowledgments in the unnumbered footnote on the first page.

	%%%%%%%%%%%%%%%%%%%%%%%%%%%%%%%%%%%%%%%%%%%%%%%%%%%%%%%%%%%%%%%%%%%%%%%%%%%%%%%%
	
	\bibliographystyle{ieeetr}
	\bibliography{sample_FG2020}
	
\end{document}